\pdfoutput=1

\documentclass[11pt]{article}

\usepackage{authblk}
\usepackage{graphics}
\usepackage{graphicx}
\usepackage{amsmath}
\usepackage{amssymb}
\usepackage{pifont}
\newcommand{\cmark}{\ding{51}}%
\newcommand{\ccmark}{\cmark \cmark}%
\newcommand{\cccmark}{\cmark \cmark \cmark}%
\newcommand{\ccccmark}{\cmark \cmark \cmark \cmark}%
\newcommand{\xmark}{\ding{55}}%
\newcommand{\xxmark}{\xmark \xmark}
\newcommand{\xxxmark}{\xmark \xmark \xmark}%
\newcommand{\xxxxmark}{\xmark \xmark \xmark \xmark}%

\usepackage[]{acl}

\usepackage{times}
\usepackage{latexsym}

\usepackage[T1]{fontenc}

\usepackage[utf8]{inputenc}

\usepackage{microtype}

\newcommand*{\affaddr}[1]{#1} 
\newcommand*{\affmark}[1][*]{\textsuperscript{#1}}
\newcommand*{\email}[1]{\texttt{#1}}

\title{Audience-Centric Natural Language Generation via Style Infusion}

\author{%
Samraj Moorjani\affmark[1], Adit Krishnan\affmark[1], Hari Sundaram\affmark[1], Ewa Maslowska\affmark[1], Aravind Sankar\affmark[1] \\
\affaddr{\affmark[1]University of Illinois at Urbana-Champaign, USA}\\
\email{\{samrajm2, aditk2, hs1, ehm, asankar3\}@illinois.edu}\\
}

\usepackage{booktabs}

\usepackage{tabularx}

\usepackage{siunitx}
\newcommand{\ie}{\textit{i.e.,~}}
\newcommand{\eg}{\textit{e.g.,~}}

\usepackage{cleveref}

\begin{document}
\maketitle

\begin{abstract}

Adopting contextually appropriate, audience-tailored linguistic styles is critical to the success of user-centric language generation systems (\eg chatbots, computer-aided writing, dialog systems). While existing approaches demonstrate textual style transfer with large volumes of parallel or non-parallel data, we argue that grounding style on audience-independent external factors is innately limiting for two reasons. First, it is difficult to collect large volumes of audience-specific stylistic data. Second, some stylistic objectives (\eg persuasiveness, memorability, empathy) are hard to define without audience feedback. 

In this paper, we propose the novel task of \textit{style infusion} - infusing the stylistic preferences of audiences in pretrained language generation models. Since humans are better at pairwise comparisons than direct scoring - \ie \textit{is Sample-A more persuasive/polite/empathic than Sample-B} - we leverage limited pairwise human judgments to bootstrap a style analysis model and augment our seed set of judgments. We then infuse the learned textual style in a GPT-2 based text generator while balancing fluency and style adoption. With quantitative and qualitative assessments, we show that our infusion approach can generate compelling stylized examples with generic text prompts. The code and data are accessible at \url{https://github.com/CrowdDynamicsLab/StyleInfusion}.

\end{abstract}

\section{Introduction}

In this paper, we develop a novel approach to infuse audience-centric styles into pretrained language generation (NLG) models. Learning to synthesize subjective styles is crucial to various applications. For instance, persuasion and memorability in computational advertising and marketing \cite{van-noort-2020-automatic}. User-centric applications of language generation, such as writing aids, chatbots, and dialog systems, often require these stylistic adjustments depending on both the audience and the task. Prior work often defines textual style with large static sentence collections. However, stylistic objectives such as persuasiveness, memorability, and empathy are hard to define without a target audience~\cite{bell1984style} due to non-uniform stylistic expectations across diverse user groups. Thus, we suggest that subjective text styles and traits must be defined by the target audience instead of audience-independent data. Our work focuses on two resulting challenges - first, how to collect target audience feedback, and second, how to leverage the limited feedback efficiently for style infusion.

Textual styles - \ie different linguistic presentations of the same conceptual content - play an integral role in persuasive/memorable communication. For instance, an informal style is less persuasive in formal settings~\cite{kim2019comparing}. The style problem extends across diverse domains, from empathic styling in mental health~\cite{cameron2018assessing} to fact-driven, simplistic styling in tech support~\cite{okuda2018ai}. Existing work in textual style transfer (TST) takes two general approaches. The strictly supervised approaches leverage fixed parallel corpora, analogous to machine translation~\cite{hu2017toward}, while semi-supervised and unsupervised techniques leverage non-parallel collections of stylized sentences~\cite{shen2017style}. Predefined metrics, heuristics, external oracles, and hybrid approaches have also been considered~\cite{jain2019unsupervised, jin2019imat}.

Constructing audience-centric or time-evolving / adaptive methods for style transfer remains an open challenge. Existing approaches are guided by rigid modeling considerations and the distributions of fixed style-specific corpora. This is innately limiting for stylistic objectives such as persuasiveness, a trait with a widely disputed definition in existing literature, and multiple external confounds such as preexisting biases and independent features of the persuader (\eg how many followers they have) \cite{al-khatib-etal-2020-exploiting,moran-2016,lowrey-1998, berger-2012,murphy-2001}. Furthermore, it is infeasible to collect extensive annotated collections of text for each audience, style, and application~\cite{pennebaker1999linguistic}. 


Unlike prior work, we define and incorporate style grounded on our target audience. To address dynamic settings requiring audience-centric linguistic styles, we propose the novel \textit{style-infusion} task. Since human reviewers are better at pairwise style comparisons than direct scoring~\cite{shah-cardinal-ordinal}, we formulate \textit{style infusion} as follows: how do we \textit{infuse} the stylistic preferences of our audience, via pairwise sentence comparisons, in a generative language model (LM)? Unlike conventional style transfer, our task leverages domain and audience-specific feedback instead of parallel non-parallel sentence collections rendered in any specific style. Further, we adopt an incremental training approach rather than retraining models from scratch.

We bootstrap an initial style analysis model to discriminate the positive and negative samples from audience feedback. Our model then selects additional samples from a generic topical sentence collection to expand the seed set of audience judgments. By separating style analysis and text generation models, we create an adversarial setup to infuse the audience's stylistic feedback in any generative LM. We weight the noisy reward from the style analysis model (discriminator) with a reconstruction loss to balance style adoption and fluency. 

In summary, our contributions are as follows: 
\begin{enumerate}
    \item \textbf{Audience-centric Style Infusion}: To our knowledge, we are the first to formulate the task of style infusion to tether the definition of style to the target audience. In contrast, prior work defines style in a purely data-driven manner~\cite{shen2017style, yang2018unsupervised}. External data limits the definition of style to the context in which it was collected. We propose a more human-centric approach to text styling through explicit audience feedback via pairwise comparisons.
    
    \item \textbf{Decoupling Style}: We decouple the style analysis and language generation models for versatility and simplicity. Prior work often unifies these tasks in a single training setup, thus sacrificing incremental learning and infusion of new stylistic preferences of audiences~\cite{jain2019unsupervised, jin2019imat}. We introduce an automatically weighted loss, combining an independent reconstruction loss for generation and discriminator-based loss for style, producing a more robust representation of style than in fused settings. 

    \item \textbf{Automatic Style Evaluation}: To the best of our knowledge, we are the first to automatically evaluate the transfer of memorability/persuasiveness. Existing literature has relied on costly manual evaluation as these two traits are hard-to-define stylistic objectives lacking generative work~\cite{li2020,tan2016,danescu-niculescu-mizil-etal-2012-hello}. We introduce a new audience-centric correlation metric using a hierarchical Bayesian model to compute the correlations of linguistic features with audience feedback. We then evaluate our model's generations based on their agreements with these audience correlations. 
\end{enumerate}

\section{Related Work}

Prior work has explored "style transfer" in diverse settings ranging from ``clickbait'' headlines to formalizing text \cite{jin2020hooks, chawla-yang-2020-semi, xu2019clickbait}. While strictly supervised approaches show high fidelity to input samples~\cite{hu2017toward, jhamtani-etal-2017-shakespearizing}, unsupervised and minimally supervised learning are widely applicable since parallel samples are unavailable \cite{shen2017style, yang2018unsupervised}.

Disentanglement, prototype editing, and pseudo-parallel corpus creation are popular approaches. Prototype editing applies stylistic markers to predefined sentence templates \cite{guu2018generating, li-etal-2018-delete}, disentanglement extracts style independent of the content~\cite{shen2017style, hu-et-al-controllable}. Audience-centric feedback may not conform to these rigid hypotheses. First, unconstrained generation allows for freedom in sentence and paragraph-level constructs to define the style \cite{li2020}. Second, the separability of content and style is harder in specialized domains reliant on domain-specific jargon~\cite{woodward2008more} and expressions. Our bootstrapping approach shares some commonalities with pseudo-parallel corpus creation (e.g. aligning sentences from two mono-style corpora) \cite{jin-etal-2019-imat, zhang2018style}, but only utilizes a generic topical corpus to expand the audience-generated ``seed set'' of pairwise judgments. Adversarial training has also been used to quantify style~\cite{yang-2022-adversarial}. Our approach explicitly decouples the style discrimination and generation tasks for modularity and incremental training purposes.

We pick two stylistic objectives that are highly audience-dependent and hard to define objectively - memorability and persuasiveness - to evaluate our approach. Prior work in these styles has been limited to analysis but not generation. \citet{tan2016} and \citet{li2020} find linguistic patterns, interaction dynamics, and discourse structure are strong identifiers of persuasive arguments, while convincingness \cite{habernal-gurevych-2016-argument}, memorability \cite{danescu-niculescu-mizil-etal-2012-hello} have been better explained by linguistic feature correlation. However, there is a lack of work on unconstrained generation of persuasive and memorable text~\cite{duerr-2021, van-noort-2020-automatic}. Our approach enables us to bridge some of these specific gaps while maintaining a generalized overall formulation.



\section{Discriminative Language Model}
\label{sec_discriminator}

In this section, we train a BERT-based style discriminator to provide feedback to our generator.

\subsection{Model Architecture and Training}

Our style discriminator (style analysis module) adds a fully connected (FC) layer with dropout to pre-trained BERT~\cite{devlin-etal-2019-bert}. We use the 'bert-base-uncased' model \cite{wolf2019} (12-layers, 768 dimension). We concatenate with a '<SEP>' token and jointly tokenize the compared pair of sentences. The FC layer generates a single output ($\mathbb{R}^{768} \rightarrow \mathbb{R}$). We threshold the sigmoid of the output at 0.5 to decide the preferred sentence. We train all layers (including BERT) on the pairwise audience feedback (batch size 32, 5 epochs, $\eta=0.0001$, dropout~$=0.2$). We also train a Siamese BERT architecture \cite{reimers2019} with the same settings but find it to underperform BERT (results in~\Cref{appendix:siamesebert}).

\subsection{Pairwise Feedback Datasets}
\label{subsection:pairwise_datasets}
We select one pairwise feedback dataset for both the persuasiveness and memorability tasks to evaluate our approach. The UKPConvArg1 corpus~\cite{habernal-gurevych-2016-argument} presents pairs of arguments where human annotators select the more persuasive argument. The authors generate 16,000 argument pairs over 16 distinct, non-overlapping topics. Both arguments in a pair belong to the same topic and argue for the same stance (\ie parallel pairwise feedback). For memorability, we leverage the Cornell Movie-Quotes Corpus~\cite{danescu-niculescu-mizil-etal-2012-hello}, containing 2,200 paired movie quotes with crowd-sourced memorability annotations. 

\subsection{Observations and Validation}
Our discriminator achieves 89\% accuracy over 5-fold cross-validation for the persuasiveness task. We further validate for overfitting by holding out two topics from the test set and training on the remaining topics, ensuring the discriminator has no exposure to these held-out topics during training. After training from scratch, the discriminator still achieves 87\% accuracy on the held-out topics. On the Cornell Movie-Quotes corpus, the discriminator achieves  80\% accuracy. We repeat the held-out topic test to validate the classification performance for the memorability task.

In summary, these tests validate the ability of our style discriminator to learn audience style preferences with small volumes of pairwise feedback. In \Cref{sec_generator}, we describe our approach to infuse the style discriminator feedback into a generative language model.

\section{Style-Aware Language Generation}
\label{sec_generator}

\begin{figure*}[ht]
  \centering
  \caption{Training diagram that shows how the loss is calculated as a weighted sum of the discriminator ($L_{D}$) and reconstruction ($L_{R}$) loss. $\alpha_S$ is decided by the discriminator as a form of contrastive learning.}
  \includegraphics[width=0.9\linewidth]{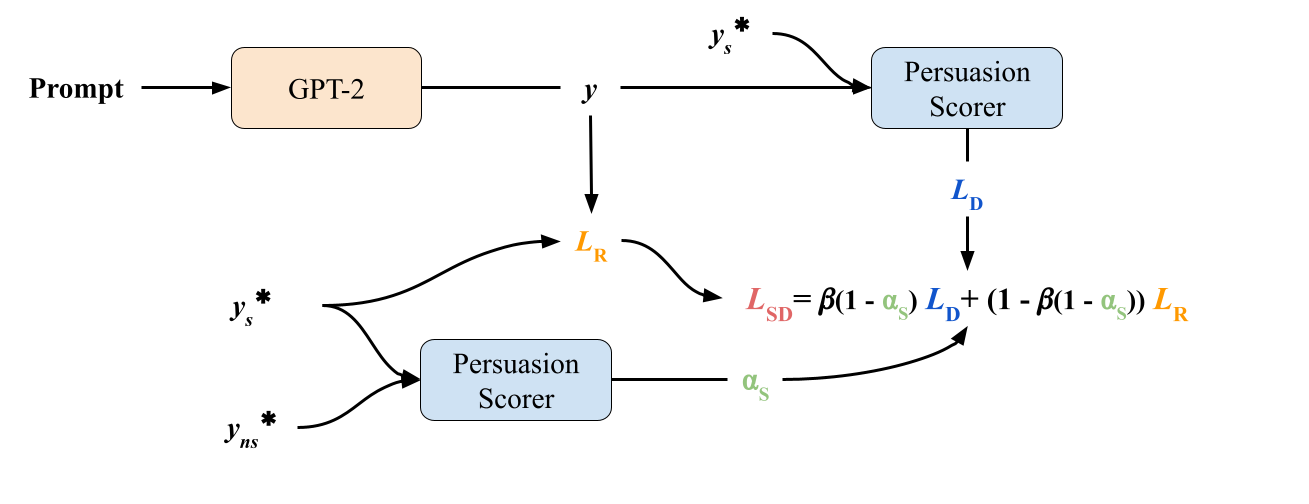}
  \label{fig:architecture}
  \vspace{-5pt}
\end{figure*}

In this section, we infuse the stylistic preferences learned by our style discriminator in \Cref{sec_discriminator} into a GPT-2 model~\cite{radford2019language} pretrained on the causal language modeling (CLM) objective~\footnote{https://huggingface.co/gpt2}. The model takes in a textual prompt and generates text, $y$, that we want to infuse with the audience preferred style. For the persuasiveness task, the UKPConvArg1 dataset provides prompts for each argument pair. For the memorability task, we use the previous sentence as the prompt.

During training, we feed the prompt and feedback pair to GPT-2 - the more preferred (styled) sample, $y^*_s$, and the corresponding less-preferred (non-styled) sample, $y^*_{ns}$. We use an adversarial training paradigm to enable the generator to learn from the discriminator, illustrated in~\Cref{fig:architecture}. 

\subsection{Training}

We utilize two losses during training: a reconstruction loss, $L_R$, and a discriminator loss, $L_D$. The reconstruction loss is meant to maximize the log probability of the styled training argument, $y_s^*$.

\begin{equation}
    L_{R} = -\frac{1}{N} \sum_{i=1}^{N} \log p(y_s^{*(i)})
\end{equation}

The reconstruction loss teaches the model to mimic the gold-standard samples. The discriminator loss is meant to maximize the score of the discriminator, $\mathcal{D}$, and is formulated as:

\begin{equation}
    L_{D} = \mathcal{D}(y_s^*, y) - \frac{1}{N} \sum_{i=1}^{N} \hat{R}_i \log p(y^{(i)})
\end{equation}

where $y^{(i)}$ is the $i$-th token of the generated sentence $y$, and $\hat{R}_i$ is a baseline reward meant to reduce the noise from the discriminator. We elaborate on the baseline reward in~\Cref{appendix:baselinereward}. 

We find that too strong of a discriminator loss negatively impacts fluency. Thus, we introduce a regularization constant, $\beta$, to ensure that the discriminator loss remains only a fraction of the loss. The two losses are weighted together to create the final loss as follows:

\begin{equation}
    L_{SD} = C \cdot L_{D} + (1-C) \cdot L_{R}
\end{equation}

where $C = \beta (1-\alpha_{S})$ and $\alpha_{S} = \mathcal{D}(y_s^*, y_{ns}^*)$. Note that $y_{ns}^*$ is the non-styled training argument. 

Instead of making the weighted ratio between the two losses constant, we make them sample dependent. The intuition is that when $\alpha_{S}$ is high (e.g. the sample is persuasive), we can just use the reconstruction loss to replicate the gold standard which will directly reflect the style. However, when $\alpha_{S}$ is low (i.e. we have a weak sample), we instead switch to learning the trends from the discriminator. This loss is referred to as the sample-dependent discriminator (SD) loss. We also compare the discriminator loss with a simpler supervised loss defined as $L_{S} = \frac{1}{N} \sum_{i=1}^{N} \mathcal{D}(y_s^*, y^{(1..i)})$.


\subsection{Dataset Augmentation Approach}

The UKPConv1 and Cornell Movie Quotes corpora we presented in~\Cref{subsection:pairwise_datasets} provide approximately 16,000 and 2,200 unique pairs for stylistic feedback; not nearly enough to train a large language model. To increase our model's breadth of knowledge, we generate additional pairwise feedback with the CNN/Daily Mail dataset \cite{see-etal-2017-get}, containing over 300,000 unique news articles. 

First, we generate the Universal Sentence Embeddings~\cite{cer-2018} of all unique sentences in our style corpora (UKPConv1, Cornell) and external corpora (CNN/Daily Mail). For each candidate sentence in the external dataset, $s_i$, we find the top-k similar sentences ($y_1 ... y_k$) in the style corpus to be augmented. We then perform pairwise comparisons $\lor_{j} \mathcal{D}(s_i, y_j) > 0.5, j \in \{1, ..., k\}$ where if the discriminator prefers the candidate external sentence ($s_i$) over \textit{any one} of the similar sentences ($y_1 ... y_k$) from the style corpus, we include the pair. Through this bootstrapped augmentation method, we ensure we have sentences that are relatively more ``styled'', as defined by our discriminator, and similar to those in our existing corpus. 

\subsection{Style-Aware Generation with GPT-2}

The OpenAI GPT-2 \cite{radford2019language} model is a large transformer-based language model pretrained on nearly 8 million web pages, allowing generalization to many domains and tasks. This is closer to the unconstrained scenarios that we wish to target with our style-infusion task. Alternate generators such as pointer-generators rely on copying~\cite{xu2019clickbait}, thus introducing more limitations in the extent of style infusion. The ability to extensively pretrain transformer-based models makes them more widely applicable for syle infusion~\cite{gururangan2020pretraining}.

In this section, we introduced the adversarial training mechanism for the style-aware language generator and the bootstrapped data augmentation method used to produce robust generations. Next, we will introduce the baselines, evaluation metrics, and training settings.


\section{Experimental Settings}

We compare our architecture against a few strong baselines:

\textit{\textbf{Pretrained GPT-2}} \cite{radford2019language} We use this pre-trained model as a representation of average text, allowing us to show shifts in style that occur due to training.

\textit{\textbf{Fine-tuning}} We fine-tune the pre-trained GPT-2 model using the reconstruction objective on the style-specific corpus only (e.g. UKPConv1).

\textit{\textbf{Fine-tuning + Data Augmentation}}  We fine-tune the pre-trained GPT-2 model using the reconstruction objective on the augmented data.

\textit{\textbf{TitleStylist}} \cite{jin2020hooks} We adapt the stylistic headline generation framework to generate stylistic text based on a prompt. \citet{jin2020hooks} utilize a Denoising Autoencoder with parameter sharing to disentangle style from content to control the style with a set of parameters.  

\textbf{Training Settings} For all GPT-2 based models, we use a base GPT-2 model from Huggingface \cite{wolf2019} (1024 dimensions, Adam optimizer, $\eta = 5e-5$). Because of the length of our text and size of our models, we utilize DeepSpeed \cite{rasley-2020-deepspeed} to distribute training over two 32GB V100s, and we train with FP16 mixed precision. We experiment with the loss parameters of $C$ and $\beta$ and discuss our findings in~\cref{section:discussion}. 

\textbf{Evaluation Metrics} We take a deeper look into the annotator labels in the UKPConvArg1 dataset and we find that some linguistic features play a significant role in the persuasiveness of text. 

We create a hierarchical Bayesian model to find the correlation between a set of collected linguistic features and the desired style. We first take the unique sentences from a dataset and compute a set of linguistic features over them. A full list of features can be found in~\Cref{appendix:lfc}.

For each linguistic feature-topic pair, we infer the correlation between the feature and the text that demonstrates the style by running a Markov Chain Monte-Carlo (MCMC) process using the No-U-Turn Sampler (NUTS). We elaborate on the calculations in~\Cref{appendix:bayesian}. Note that the results we show are in the logit scale, meaning even a change of $\mp 1$ has a big effect on the probability (about a 23\% difference in odds of winning).  

The models then generate text based on the prompts in a held-out test set and we calculate the features of the generations. We run a t-test to determine if the difference in features between a pretrained GPT-2 and one of our models is statistically significant. This evaluation shows how our trained model learns to use these linguistic features to construct more stylized arguments.

In addition, we use \textit{pyrouge} library to collect the ROUGE \cite{lin-2004-rouge} score, a commonly used metric that measures the N-gram overlap between the training and generated arguments. While these scores will not tell us how persuasive our generations are, they will ensure that the generations remain on topic.

Lastly, we compute the BERTScore \cite{zhang2019bertscore}, another automatic evaluation metric that computes token similarity using contextual embeddings. The BERTScore represents the semantic similarity of the generations to the test set which will ensure generations are relevant, but not necessarily persuasive.

\section{Results on Persuasiveness}

In this section, we analyze our results by showing a significant usage of linguistic features that resemble persuasive text, showing generated text, and with standard metrics.  

\subsection{Linguistic Feature Correlations}
\begin{figure}[h]
  \centering
  \scalebox{1.0}{
  \includegraphics[width=\linewidth]{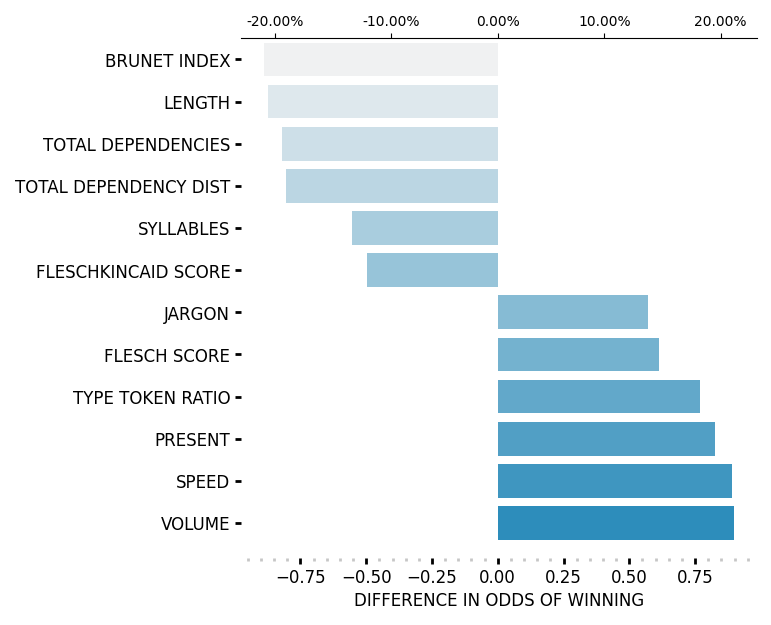}
  }
  \caption{The correlations between linguistic features and convincingness in the UKPConvArg1 corpus. The lower x-axis is in the logit scale, and the percentage difference in odds of winning is on the upper x-axis. The figure is read as: if the feature for argument A is one standard deviation greater than the feature for argument B, the odds of A winning shift by the respective percent value. Notice that the correlations show a strong positive correlation with readability (\eg the Flesh score is positive while length and average syllables are negative). }
  \label{fig:feature_correlations}
\end{figure}

\Cref{fig:feature_correlations} shows the correlations between linguistic features and convincingness in the UKPConvArg1 corpus. The model details are in~\Cref{appendix:lfc}.

We find a strong positive correlation between readability and winning arguments. This is reflected by both readability scores (\eg SMOG, Flesch-Kincaid, etc.) and correlation with smaller words, fewer total dependencies, and a smaller overall total dependency distance. We notice a positive correlation with speed and volume. \citet{toubia-2021} define speed as the total distance covered by a text's word embeddings, normalized by the length of the text. Volume represents the amount of material covered by the text, calculated by estimating the volume enclosed by the word embeddings \cite{toubia-2021}. We also find a negative correlation with passive voice and a positive correlation with misspelled words (not shown for brevity).

We run a significance test to see how well our models learned the style (see~\Cref{appendix:featureagreement}). Most models consistently learn pronounced trends (\ie Brunet index, length, speed, and volume). The augmented data likely led to this change because fine-tuning on the augmented set displays the same trends. In cases like total dependencies (TD) and the ratio of present tense verbs, models trained with the sample-dependent discriminator (SD) loss are significantly better at learning the trend, despite the data not actively showing the trend (or completely opposing it). In the case of Flesch score, models trained with SD loss can nullify the trend which occurs in the incorrect direction. This displays that models trained with the SD loss are substantially better at learning from the dataset than the baselines and models trained with the sample-dependent supervised (SS) loss. One example of failure is the ratio of jargon, likely because the model could not generate out of vocabulary words, but this is a limitation of how we define jargon.


\subsection{Sample Generations}


  \begin{table}[htb]
  \small
  \centering
  \caption{The baseline models and our model (GPT2 with the SD loss and $\beta=0.5$ on the augmented data). The models are fed the input ``Plastic bottles are bad because'' and ``Gaming is good for child development because'', generating up to 100 tokens using beam search.}
  \label{tab:example_persuasive} 
  \begin{tabularx}{\linewidth}{@{}>{\raggedright\arraybackslash}X@{}}
   \toprule[1.5pt]
  \textsc{Model} \& \textsc{Generated Text}\\
  \midrule[0.75pt]
  \textsc{Example 1:}\\
  \textbf{GPT2}:  Plastic water bottles are bad because they are not recyclable," he said. "They are not good for the environment, they are not recyclable."  \\
  \textbf{GPT2-Aug}: Plastic water bottles are bad because it causes pollution. Also water bottles also cost more than regular water. \\
  \textbf{TStylist}: Plastic water bottles are bad because plastic water bottles are bad for our environment.\\
  \textbf{Ours (SD-0.5)}: Plastic water bottles are bad because they are not recycled and end up in landfills which lead to pollution of our environment. Plastic requires up to 47 million gallons of oil per year to produce.  \\
  \addlinespace[0.5em]
  \textsc{Example 2:}\\
  \textbf{GPT2}: Gaming is good for child development because it teaches them how to use a computer. It's good for the economy, it's good for the environment, it's good for the children, it's ...  \\
  \textbf{GPT2-Aug}: Gaming is good for child development because it allows the child to grow and develop. I believe eSports (LoL) should be a spectator sport and not a major part of the sports calendar.   \\
  \textbf{TStylist}: Gaming is good for child development because gaming is good for your child.\\
  \textbf{Ours (SD-0.5)}: Gaming is good for child development because it allows children to grow up in a world where they are exposed to a wide variety of ideas and experiences.   \\
  \bottomrule[1.5pt]\\
  \end{tabularx}
  \vspace{-10px}
  \end{table}

\Cref{tab:example_persuasive} shows the generations of three baselines and our best-performing method. We find that for both prompts, the generations of models trained with the sample-dependent discriminator (SD) loss generally have the highest values of speed, volume, and lexical diversity. For the second prompt, the speed and volume of our generation are larger than that of GPT2 and TStylist, but slightly smaller than that of GPT2 fine-tuned on the augmented data. Intuitively, this makes sense because the GPT-Aug generation covers much more information in the same time frame; however, this information isn't relevant to the argument, making our generation much more sensible. The baselines often suffer from neural degeneration, but the model trained with the SD loss does not face this issue. Since length had a strong negative correlation with persuasiveness, the model likely implicitly learned from the discriminator to handle this kind of neural degeneration. However, it is still an issue in some cases with out-of-domain samples.

  \begin{table}[htb]
  
  \centering
  \small
  \caption{ROUGE-\{1,2, L\} scores and BERT scores (F1) for all models. Baseline models: GPT2, GPT-2 fine-tuned on UKPConvArg1, GPT-2 with augmented data, TitleStylist \cite{jin2020hooks}. Our models are trained on augmented data and a sample-dependent discriminator (SD) or sample-dependent supervised (SS) loss with parameter $\beta$. The baseline ROUGE score increases due to data augmentation; the relevance of our models' generations is largely insensitive to loss type and parameter value.}
  \label{tab:automatic-evaluation}
  \scalebox{1.0}{
  \begin{tabularx}{0.9\linewidth}{@{}>{\raggedright\arraybackslash}Xcccc@{}}
   \toprule[1.5pt]
  \textsc{Model} & \textsc{RG-1}            & \textsc{RG-2}           & \textsc{RG-L}        &\textsc{B-F1} \\     
  \midrule[0.75pt]
  GPT2     & 0.1856          & 0.0968          & 0.1769         & 89.28          \\
  GPT2-UKP & 0.2474          & 0.1061          & 0.1989         & 87.21         \\
  GPT2-Aug & 0.2987          & 0.1845          & 0.2774         & 86.90          \\ 
  TStylist & 0.2578          & 0.1569          & 0.2391         & 84.70          \\ 
  \midrule[0.75pt]
  SS-0.1   & 0.2925          & 0.1802          & 0.2717         & 88.95          \\
  SS-1.0   & 0.2862          & 0.1774          & 0.2634         & 89.18          \\
  SD-0.1  & 0.3036          & 0.1903          & 0.2802          & 88.75         \\
  SD-0.5  &    \textbf{0.3168} & \textbf{0.2296} & \textbf{0.3065 } & \textbf{89.56}       \\
  SD-0.8  & 0.2872          & 0.1957          & 0.2733          & 89.12         \\
  SD-1.0  & 0.2929          & 0.2224          & 0.2848          & 89.05         \\ 
  \bottomrule[1.5pt]\\
  \end{tabularx}
  }
  \vspace{-25pt}
  \end{table}

\subsection{Automatic Metrics}
We compare the ROUGE scores of our experimental models in ~\Cref{tab:automatic-evaluation}, ensuring that the topics in the test set are not discussed anywhere in the UKPConvArg1 or augmented datasets. The data augmentation leads to a sharp increase in the ROUGE scores of the generations, showing that it is essential for robust and relevant generations. The results are relatively insensitive to variation in $\beta$ parameter that controls the tradeoff between reconstruction loss ($L_R)$ and discriminator loss ($L_R$). These scores show that our models generate relevant, but not necessarily persuasive, text. We find similar insights from the BERTScore; although the augmentation has a slight negative impact on the score, the difference is negligible. 

\section{Results on Memorability}

In this section, we focus on memorability and show that our model can generate more robust, relevant, and memorable text than the baselines.

\subsection{Linguistic Feature Correlations}
\begin{figure}[h]
    \caption{The correlations between linguistic features and persuasiveness in the Cornell Movie-Quotes Corpus. The lower x-axis is in the logit scale, and the percentage difference in odds of winning is on the upper x-axis. Notice that the correlations show a negative correlation with long and winding text (\ie circuitousness \cite{toubia-2021}). }
  \centering
  \scalebox{1.0}{
  \includegraphics[width=\linewidth]{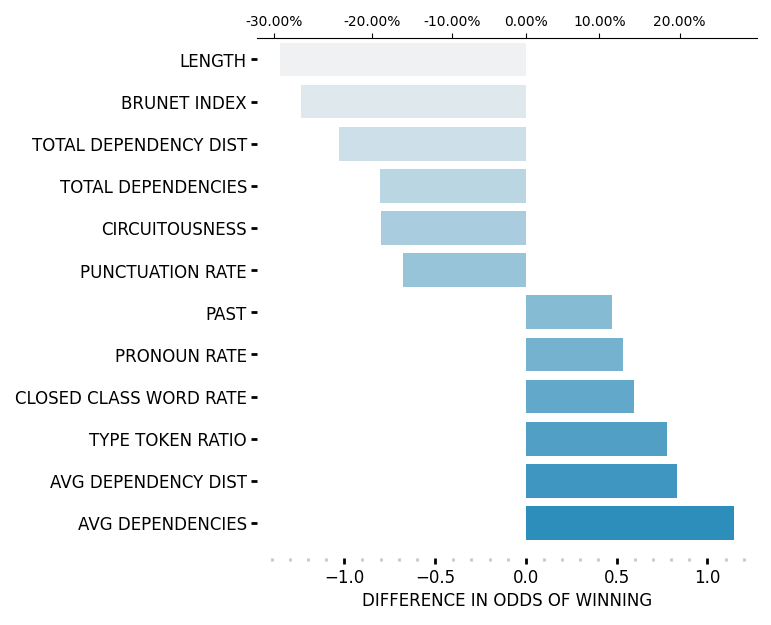}
  }

  \label{fig:mem_feature_correlations}

\end{figure}

We train a Bayesian hierarchical model for the Cornell Movie Quotes corpus, which produces the correlations shown in~\Cref{fig:mem_feature_correlations}. We find a strong negative correlation with long, winding text, shown by the trends in total dependencies, total dependency distance, length, and circuitousness. A higher circuitousness implies that a less direct route was taken to convey information \cite{toubia-2021}. Circuitousness is detrimental to memorability as winding text tends to be harder to remember. The negative correlation with the punctuation rate and positive correlation with the average dependencies show that more memorable text tends to have a few sentences, independent of the length of sentences. Lastly, there is a strong emphasis on uncommon vocabulary with a negative correlation with the Brunet index and a positive correlation with token type ratio. This is supported by findings from \citep{danescu-niculescu-mizil-etal-2012-hello} who find that memorable quotes are built upon less common word choices. 

We run the significance test on a held-out test set to see how well our models learned to generate memorable text. In some cases in~\Cref{tab:mem-feature-evaluation}, models trained with the sample-dependent discriminator (SD) loss have similar performance as the fine-tuned models, indicating that some relevant features are learned solely from fine-tuning. However, many other incorrect trends are corrected with training using the SD loss. The only feature that does not improve is the pronoun rate, likely because of shorter sentences with more emphasis on uncommon word choices. 

\subsection{Sample Generations}

We look at a few examples of generations to see how training influenced the model's generations in~\Cref{tab:example_the_more}. There is a visible decrease in the number of sentences, overall length, and circuitousness in both the GPT2-Aug and SD-0.5 models. This reflects the trends shown in~\Cref{tab:mem-feature-evaluation}, but we can see that the model trained with the SD loss generates sentences that are more sophisticated than the fine-tuned GPT2 model's. TitleStylist generates sentences that are identical to our generation feature-wise but are not as sophisticated as our generations.

  \begin{table}[htb]
  \small
  \centering
  
  \begin{tabularx}{\linewidth}{@{}>{\raggedright\arraybackslash}X@{}}
   \toprule[1.5pt]
  \textsc{Model} \& \textsc{Generated Text}\\
  \midrule[0.75pt]
  \textsc{Example 1:}\\
  \textbf{GPT2}:  The more you know about it, the more likely you are to believe that it is true. If you don't believe that it is true, you're not alone. If you don't believe that it is true, you're not alone... \\
  \textbf{GPT2-Aug}: The more you smoke, the more you will smoke  \\
  \textbf{TStylist}: The more you need to know. \\
  \textbf{Ours (SD-0.5)}: The more you learn, the more you see.  \\
  \addlinespace[0.5em]
  \textsc{Example 2:}\\
  \textbf{GPT2}: When solving problems, dig a hole in the ground and dig a hole in the ground and dig a hole in the ground and dig a hole in the ground and ...    \\
  \textbf{GPT2-Aug}: When solving problems, dig your heels in and try to find a solution, even if you don't have the answers, and even if you don't know the answers. \\
  \textbf{TStylist}: When solving problems, dig better and better.\\
  \textbf{Ours (SD-0.5)}: When solving problems, dig deeper than a grave.   \\
  \bottomrule[1.5pt]
  \caption{Generations of GPT2, GPT2 fine-tuned on augmented data, and GPT2 with the SD loss ($\beta=0.5$) on the augmented data. The models are fed in the inputs ``The more you'' and ``When solving problems, dig''.}
  \label{tab:example_the_more}
  \vspace{-15pt}
  \end{tabularx}
  \end{table}

 \subsection{Automatic Metrics}
Similar to persuasiveness,~\Cref{tab:mem-automatic-evaluation} shows that the ROUGE scores increase mainly due to data augmentation. Once again, these results demonstrate that the augmented data leads to more relevant generations, increasing the breadth of knowledge transferred to the model. The same trends generally hold for the BERTScore, which shows that the generations remain semantically relevant.


\begin{table}[t]
  \centering
  \small
  \caption{ROUGE-\{1,2, L\} scores and BERT scores (F1) for all models. Baseline models: GPT2, GPT-2 fine-tuned on UKPConvArg1, GPT-2 with augmented data, TitleStylist \cite{jin2020hooks}. Our models are trained on augmented data, and a sample-dependent discriminator (SD) or sample-dependent supervised (SS) loss with parameter $\beta$. The baseline ROUGE score increases due to data augmentation; again, the relevance of generations is largely independent of loss type and parameter value.}
  \label{tab:mem-automatic-evaluation}
  \begin{tabularx}{0.9\linewidth}{@{}>{\raggedright\arraybackslash}Xcccc@{}}
   \toprule[1.5pt]
  \textsc{Model} & \textsc{RG-1}            & \textsc{RG-2}           & \textsc{RG-L}    & \textsc{B-F1} \\     
  \midrule[0.75pt]
GPT2      & 0.1503          & 0.0853          & 0.1461          & 81.12          \\
GPT2-IMDB & 0.1579          & 0.0853          & 0.1510          & \textbf{88.87}          \\
GPT2-Aug  & 0.2737          & 0.1703          & 0.2685          & 87.24          \\
TStylist  & 0.2542          & 0.1617          & 0.2439          & 85.99          \\
  \midrule[0.75pt]
  SS-0.1  & 0.2746          & \textbf{0.1759} & 0.2668          & 85.98          \\
  SS-1.0  & 0.2740          & 0.1723          & 0.2661          & 85.93         \\
  SD-0.1  & 0.2743          & 0.1735          & 0.2686          & 86.87          \\
  SD-0.5  & 0.2718          & 0.1706          & 0.2680          & 83.94         \\
  SD-0.8  & \textbf{0.2812} & 0.1705          & \textbf{0.2733}    & 85.81       \\
  SD-1.0  & 0.2739          & 0.1681          & 0.2675          & 86.12         \\
  \bottomrule[1.5pt]\\
  \end{tabularx}
  \vspace{-10pt}
  \end{table}

We show that our model generates more robust, relevant, and memorable text than the baselines. Next, we discuss how tuning the loss parameters affects generations.

\section{Empirical Observations}
\label{section:discussion}

We analyze how the value of $\beta$ affects generations, finding that generations from $\beta=0.1$ suffer the same degeneration as fine-tuning while higher values avoid these issues. Because $\alpha_{S}$ is not always 1, the constant in front of the discriminator loss is less than $\beta$. Consequently, the discriminator is not given enough weight, and the generator cannot learn as effectively from the discriminator. It is difficult to distinguish differences between $\beta=0.5$ and $\beta=1.0$, but aside from $\beta=0.5$, $\beta=1.0$ outperforms every other value of $\beta$.  

\begin{table}[t]
  \small
  \centering
  \caption{Generations of GPT2 trained with the sample-dependent discriminator loss objective with different values of $\beta$. The generations for SD-0.5 and SD-1.0 tend to be much better than for SD-0.1}
  \label{tab:example_beta}
  \begin{tabularx}{\linewidth}{@{}>{\raggedright\arraybackslash}X@{}}
   \toprule[1.5pt]
  \textsc{Model} \& \textsc{Generated Text}\\
  \midrule[0.75pt]
  \textbf{SD-0.1:} IE sucks and makes development on your computer much more difficult than it should be. I believe that Internet Explorer (IE) is far inferior to Internet Explorer (IE) and Internet Explorer (IE) is far inferior to Internet Explorer (IE) \\
\textbf{SD-0.5:} IE sucks and makes development more difficult \\
\textbf{SD-1.0:} IE sucks and makes development on your computer much more difficult than it should be.  \\

  \bottomrule[1.5pt]\\
  \end{tabularx}
  \vspace{-20pt}
  \end{table}

We also experiment with hard-coding the coefficients for the discriminator and reconstruction loss in~\Cref{tab:example_hardcoded_weights}. Putting too much weight on the discriminator loss, $L_D$, (\ie 0.9) leads to poor quality arguments having some of the strongest linguistic feature changes (\eg shorter length). Conversely, limiting $L_D$ to 0.1 leads to much stronger generations. We introduced the $\beta$ parameter to cap $L_D$ at $\beta$. Because of the $\beta$ parameter, the previous experiments show similar but less obvious trends. 


\begin{table}[t]
  \small
  \centering
  \caption{Generations of our model trained with a mixed reconstruction and discriminator loss objective with hard-coded weights (as opposed to sample-dependent). }
  \label{tab:example_hardcoded_weights}
  \begin{tabularx}{\linewidth}{@{}>{\raggedright\arraybackslash}X@{}}
   \toprule[1.5pt]
  \textsc{Model} \& \textsc{Generated Text}\\
  \midrule[0.75pt]
  \textsc{Example 1:}\\
  
  \textbf{0.9 Supervised + 0.1 MLE:} Schools should teach  \\
physical education because it's a good thing. \\
\textbf{0.1 Supervised + 0.9 MLE:} Schools should teach  \\
physical education because PE helps children develop  \\
good habits later on in life. Plus, there's the benefit of \\
working together as a team that doesn't always happen in \\
other classes. \\

  \addlinespace[0.5em]
  \textsc{Example 2:}\\

\textbf{0.9 Supervised + 0.1 MLE:} Plastic water bottles \\
are bad because they are not recyclable. \\
\textbf{0.1 Supervised + 0.9 MLE:} Plastic water bottles \\
are bad because they are bad for the environment and they \\
are bad for the economy.Some people think that bottled \\
water is bad for consumers and should only be used in \\
situations such as disasters when no other clean water \\
is available.\\ 

  \addlinespace[0.5em]
  \textsc{Example 3:}\\

\textbf{0.9 Supervised + 0.1 MLE:} Gaming is good for \\
child development because you can play with other kids. \\
\textbf{0.1 Supervised + 0.9 MLE:} Gaming is good for child \\
development because it teaches them how to think and  \\
solve problems. It also teaches them how to communicate   \\
with each other. \\

  \bottomrule[1.5pt]\\
  \end{tabularx}
  \vspace{-20pt}
  \end{table}

\section{Conclusion}
In this paper, we introduced \textit{style infusion} to motivate infusing audience-centric, stylistic preferences into unconstrained natural language generation models. We present a bootstrapped data augmentation method for limited pair-wise audience feedback and an adversarial training framework with a decoupling loss to train a style-infused GPT-2. Through an automatic evaluation method for the transfer of audience-specific styles, we show that our approach generates compelling stylized examples with generic text prompts better than the baselines. 

Synthesizing text with subjective styles, such as persuasion and memorability, remains a significant challenge in domains like computational advertising. Our work takes the first few steps to address this problem. We plan to continue improving our work in many directions, such as incorporating long-document attention mechanisms \cite{beltagy2020longformer} to capture document-level style features and altering the discourse structure to convey information in a more interpretable manner. 

\newpage

\section{Limitations}
As with other unconstrained natural language generation applications, our system is prone to issues like degeneration from beam search and neural hallucinations. To combat the former, we post-process generations, but future work will hopefully provide better methods to prevent this issue. For the latter, we increase our dataset with samples from the CNN/DM dataset, partially mitigating the problem, but out-of-domain topics still suffer. Increasing the dataset size will only work for so long due to diminishing marginal returns.

Due to the limited amount of data available, we considered iteratively training the discriminator with the augmented data while we trained the generator. Ultimately, we felt that the weak labels would dilute the learned trends in the discriminator, but it may be interesting to see how it affects the performance of the framework. Currently, collecting pairwise datasets to use with this framework can be viewed as a limitation. With increasing interest in the computational synthesis of persuasive text and imagery, we expect to see more relevant curated datasets in the near future. Generating pairwise data through human subject experiments is expensive, which is why the data augmentation methods introduced in this paper are crucial for future work. 

We also note that our framework is limited by the computational resources available to us. Thus, we were unable to effectively support long text generation while preserving the quality of the generated text. During training, we decrease the batch size and utilize the DeepSpeed framework \cite{rasley-2020-deepspeed}, but it is still insufficient to handle long text. Furthermore, traditional left-to-right generation struggles with long text as the topics tend to diverge. Because many styles, like persuasiveness, are dependent on paragraph-level features in addition to sentence-level ones, it is beneficial for our application to support longer texts.

Lastly, one of the biggest limitations of this paper is in showing the effectiveness of the architecture we choose. Because most baselines are in style transfer and fundamentally differ from our task, we find it difficult to make a fair comparison with prior work. Regardless, style infusion is a critical step for unconstrained NLG systems such as dialogue systems and chatbots, especially in the context of human-centric stylistic objectives, which are already difficult enough to define.

\section{Ethics Statement and Broader Impact}

Our objective for developing a stylistic generative language model that leverages domain and audience-specific feedback is to enable unconstrained generation applications to appeal to more human users. For example, generating more persuasive real news might help combat misinformation by propagating the truth faster than falsehoods. In advertising and communication, persuasiveness and memorability are critical traits and having an unconstrained generation model that could replicate these features would have a multitude of positive applications, especially in targeted interventions. Previous research has mostly focused predicting audience characteristics and targeting, but not on synthesizing matching messages.

We acknowledge the dual-use concerns of the misuse of such a generation framework to, for example, spread misinformation. For this reason, we do not release the model or the pretrained generator checkpoint used in this work. 

\section*{Acknowledgements}
This work used the Extreme Science and Engineering Discovery Environment (XSEDE) Expanse GPU cluster, which is supported by National Science Foundation grant number ACI-1548562~\cite{xsede}. This work was also supported by the National Center for Supercomputing Application's Nano cluster.

\bibliography{custom}
\bibliographystyle{acl_natbib}

\appendix

\label{sec:appendix}

\begin{table*}[hbt]
\scalebox{0.7}{
\begin{tabular}{|l|cccccc|cccccc|}
\hline
Model    & BI                                            & Length                                        & TD                                            & TDD                                           & Syllables           & Flesch-Kincaid                        & Jargon       & Flesch                                        & TTR                                   & Present                                       & Speed                                 & Volume                                \\ \hline
GPT2     & -                                             & -                                             & -                                             & -                                             & -                   & -                                     & -            & -                                             & -                                     & -                                             & -                                     & -                                     \\
GPT2-16k & \cmark                                  & \xxxxmark         & \xxxxmark         & -                                             & -                   & -                                     & \xmark & -                                             & \ccmark                   & \xxxxmark & -                                     & \ccccmark \\
GPT2-Aug & \ccccmark & \ccccmark & \cmark                                  & \ccccmark & -                   & \xxxxmark & -            & \xxxxmark & \cmark                            & -                                             & \ccccmark & \ccmark                   \\
TStylist & \cmark & \ccccmark & \cccmark                                  & - & \ccmark                   & \xxxxmark & \xxxxmark            & \xxxxmark & \cmark                            & -                                             & \cccmark & \cccmark                   \\ \hline
AP-0.1   & \ccccmark & \ccccmark & \ccmark                       & \ccccmark & -                   & \xxxxmark & -            & \xxxxmark & \ccmark                   & \cmark                                    & \ccmark                   & \cccmark          \\
AP-1.0   & \ccccmark & \ccccmark & \cmark                                  & \ccccmark & -                   & \xxxxmark & -            & \xxxxmark & \ccmark                   & -                                             & \cccmark          & \ccccmark \\
SD-0.1  & -                                             & -                                             & \cmark                                  & -                                             & \xxmark & \xxxxmark & -            & \xxxxmark & \xmark                          & -                                             & \cmark                            & -                                     \\
SD-0.5  & \ccccmark & \ccccmark & \ccccmark & \ccccmark & -                   & -                                     & -            & \xmark                                  & \ccccmark & \ccccmark         & \ccccmark & \ccccmark \\
SD-0.8  & -                                             & \ccccmark & \ccccmark & \cmark                                  & -                   & \xxmark                   & -            & \xxxmark            & \ccmark                   & \cmark                                    & \ccccmark & \ccccmark \\
SD-1.0  & \ccccmark & \ccccmark & \ccccmark & \cccmark            & \xmark          & -                                     & -            & -                                             & \ccccmark & \ccccmark         & \ccccmark & \ccccmark \\ \hline
\end{tabular}}
\caption{\label{tab:feature-evaluation}Significance tests on the change in features between a pretrained GPT-2 model and a trained model. In order, these features are: Brunet Index (BI), Length (in characters), Total Dependencies (TD), Total Dependency Distance (TDD), Average Syllables per Word, Flesh-Kincaid readability score, Ratio of Jargon (i.e. out of vocab words), Flesch readability score, Token Type Ratio (TTR), Ratio of Present Tense Verbs, Speed, and Volume. In the table, the number of checks or crosses indicates the level of p-value and the correctness of the direction of the trend. Note that \cmark : $p < 0.05$, \ccmark : $p < 0.01$, \cccmark : $p < 0.001$, \ccccmark : $p < 0.0001$.}
\end{table*}

\begin{table*}[hbt]
\scalebox{0.65}{
\begin{tabular}{|l|cccccc|cccccc|}
\hline
Model    & Length                                            & BI                                        & TDD                                            & TD                                           & Circuitousness           & PunctRate                        & Past       & Pronoun Rate                                        & CCW Rate                                   & TTR                                       & ADD                                 & AD                                \\ \hline
GPT2     & -                                             & -                                             & -                                             & -                                             & -                   & -                                     & -            & -                                             & -                                     & -                                             & -                                     & -                                     \\
GPT2-IMDB & - & \ccccmark                                 & \xxmark & - & \xmark                   & \cmark & \cccmark            & \xxxxmark & \ccccmark                            & -                                             & \ccccmark & \ccccmark                   \\
GPT2-Aug & \ccccmark & \ccccmark                                  & \ccccmark & \ccccmark & \ccccmark                   & \cmark & \ccccmark            & \xxxxmark & \xxxxmark                            & \ccccmark                                             & \xxxxmark & \xxxxmark                   \\ 
TStylist & \ccccmark & \ccccmark & \cccmark & \cccmark & \cccmark & \cmark & \ccccmark & \xxxmark & - & \ccccmark & \xxxmark & \ccccmark \\\hline
AP-0.1 & \ccccmark & \ccccmark & \ccccmark & \ccccmark & \ccccmark & - & \ccccmark & \xxxxmark & \xxxxmark & \ccccmark & \xxxxmark & \xxxxmark  \\
AP-1.0 & \ccccmark & \ccccmark & \ccccmark & \ccccmark & \ccccmark & - & \ccccmark & \xxxxmark & \xxxxmark & \ccccmark & \xxxxmark & \xxxxmark  \\
SD-0.1 & \ccccmark & \ccccmark & \ccccmark & \ccccmark & \ccccmark & - & \ccccmark & \xxxxmark & \xxxmark & \ccccmark & \xxxxmark & \xxxxmark  \\
SD-0.5   & \ccccmark & \ccccmark                                  & \ccccmark & \ccccmark & \ccccmark                   & \cmark & \ccccmark            & \xxxmark & \ccccmark                            & \ccccmark                                             & \ccmark & \ccccmark  \\
SD-0.8 & \ccccmark & \ccccmark & \ccccmark & \ccccmark & \ccccmark & - & \ccccmark & \xxxxmark & \xxmark & \ccccmark & \xxxxmark & \xxxxmark  \\
SD-1.0  & \ccccmark & \ccccmark                                  & \ccccmark & \ccccmark & \ccccmark                   & \ccmark & \ccccmark            & \xxxxmark & \cccmark                            & \ccccmark                                             & \cmark & \ccccmark  \\ \hline
\end{tabular}}
\caption{\label{tab:mem-feature-evaluation}Significance tests on the change in features between a pretrained GPT-2 model and a trained model. In order, these features are: Length (in characters), Brunet Index (BI), Total Dependency Distance (TDD), Total Dependencies (TD), Circuitousness, Punctuation Rate, Past, Pronoun Rate, Closed Class Word (CCW) Rate, Token Type (TTR) Ratio, Average Dependency Distance (ADD), Average Dependencies (AD). In the table, the number of checks or crosses indicates the level of p-value and the correctness of the direction of the trend. Note that \cmark : $p < 0.05$, \ccmark : $p < 0.01$, \cccmark : $p < 0.001$, \ccccmark : $p < 0.0001$.}
\end{table*}

\section{Siamese BERT Discriminator}
\label{appendix:siamesebert}

To validate our results, we tried another architecture, similar to Siamese BERT \cite{reimers2019}, where we tokenized the texts individually and passed them through their own BERT layers, producing two embeddings, $e_1$ and $e_2$. We concatenated the two outputs along with the distance between the two as follows:
\begin{equation*}
    [e_1 ; e_2 ; |e_2 - e_1|]
\end{equation*}
We passed this new vector of $\mathbb{R}^{3 \times h}$, where $h$ is the hidden dimension of BERT, through a fully connected classification layer.

While the original discriminator achieves approximately 89\% accuracy on the random test set, the Siamese BERT model achieves a smaller, but still significant, 83\% accuracy. On the Cornell Movie-Quotes corpus, with the same hyperparameters, the original discriminator achieves 80\% accuracy and a 77\% accuracy with the Siamese BERT architecture. We choose to use the simpler discriminator architecture because it seems to capture style better than the Siamese BERT architecture. 

  \begin{table*}[htb]
  \centering
  \caption{Percentage agreement with the linguistic feature correlations calculated using the hierarchical Bayesian model. Baseline models: GPT2, GPT-2 fine-tuned on UKPConvArg1 or the Cornell Movie Quotes corpus, GPT-2 with augmented data, and TitleStylist \cite{jin2020hooks}. Our models are trained on augmented data and a sample-dependent discriminator (SD) or sample-dependent supervised (SS) loss with parameter $\beta$. We show that our models are significantly better at learning stylistic features compared to our baselines.}
  \label{tab:agreement-evaluation}
  \scalebox{0.8}{
  \begin{tabularx}{0.6\linewidth}{@{}>{\raggedright\arraybackslash}Xccc@{}}
   \toprule[1.5pt]
  \textsc{Model} & \textsc{Persuasiveness}            & \textsc{Memorability}   \\     
  \midrule[0.75pt]
GPT2     & 29.51          & 40.71                 \\
GPT2-FT  & 41.03          & 46.22                  \\
GPT2-Aug & 44.01          & 51.26                \\
TStylist & 35.76          & 43.86                   \\
  \midrule[0.75pt]
SS-0.1   & 42.26          & 49.70                \\
SS-1.0   & 48.57          & 44.97                   \\
SD-0.1  & 43.58          & 48.73                 \\
SD-0.5  & 48.56          & \textbf{62.35} \\
SD-0.8  & 50.04          & 48.14              \\
SD-1.0  & \textbf{50.18}          & 55.61                   \\
  \bottomrule[1.5pt]\\
  \end{tabularx}}
  \end{table*}

\section{Baseline Reward}
\label{appendix:baselinereward}

The baseline reward is meant to reduce the noise from the reward given by our discriminator \cite{ranzato2015sequence}. The baseline reward, $\hat{R}_i$, is calculated using a linear layer, with the input being the hidden states of our generator at timestep $i$. The intuition is that the linear layer approximates the value of the reward for a certain timestep and in practice, reduces the variance from the reward. We train the linear layer with the following loss:

\begin{equation}
    L_{BR} = \frac{1}{N} \sum_{i=1}^{N} |\mathcal{D}(y_s^*, y) - \hat{R}_i|^2
\end{equation}

where $\mathcal{D}(y_s^*, y)$ is the output of the discriminator when fed a gold argument and the generated argument (i.e. the reward). 

\section{Linguistic Feature Correlation}
\label{appendix:lfc}

\subsection{Collected Linguistic Features}

We collected the following linguistic features: length, verb tenses (e.g. future, past, etc.), punctuation rates, readability scores (Flesch score, Flesch-Kincaid score, Gunning Fog score, SMOG score, Dale-Chall score), part of speech rates (noun rate, verb rate, demonstrative rate, adjective rate, adposition rate, adverb rate, auxiliary rate, conjunction rate, determiner rate, interjection rate, numeral rate, particle rate, pronoun rate, proper noun rate, punctuation rate, subordinating conjunction rate, symbol rate, possessive rate), ratios of part of speech (e.g. noun-verb ratio, noun ratio, pronoun-noun ratio, closed-class word rate, open-class word rate), dependency information (total dependency distance, average dependency distance, total dependencies, average dependencies),  content density, idea density, lexical diversity statistics (Honore statistic, Brunet index), type token ratio, average word length, proportion of inflected verbs, proportion of auxiliary verbs, proportion of gerund verbs, proportion of participles, proportion of mispelled words, amount of alliteration, passive voice, average number of syllables, proportion of jargon, proportion of MTCG verbs (Modal, Tentative, Certainty, Generalizing), rates of named-entity recognition (NER) tags (e.g. PERSON, DATE, CARDINAL, WORK OF ART, NORP, Certainty, GPE, ORG, LOC, PERCENT, MONEY, QUANTITY, TIME, PRODUCT, EVENT, LANGUAGE, FAC), and word embedding-based measures (i.e. speed, volume, circuitousness) \citep{toubia-2021}. The NER tags were obtained from the spaCy library \citep{spacy2} and many of the linguistic features are obtained from the blabla library \citep{shivkumar2020blabla}.

\subsection{Bayesian Model}
\label{appendix:bayesian}

We define the hierarchical Bayesian model as a binomial distribution around $p$. Note that text A always demonstrates the style more strongly than text B or equally to text B. We calculate $p$ as follows:

\begin{equation}
    logit_p = \bar{p} + (\alpha[A_{id}] - \beta[B_{id}]) + \gamma[t] * (A_{ft} - B_{ft})
\end{equation}

where $\bar{p}$ is the intercept, $\alpha$ and $\beta$ are meant to capture any existing bias towards either text and $\gamma$ measures the correlation between the linguistic feature and the style for a specific topic $t$. We construct $\alpha$ and $\beta$ for all texts, hence why we index them with $A_{id}$ and $B_{id}$, respectively. Similarly, we construct $\gamma$ for each topic. $A_{ft}$ and $B_{ft}$ are the features of text A and B, respectively. $\alpha$, $\beta$, and $\gamma$ are all constructed similarly. Let's take $\alpha$ as an example:

\begin{equation}
    \alpha = \bar{\alpha} + \alpha_{v} \alpha_{\sigma}
\end{equation}

where $\bar{\alpha} \sim \mathcal{N}(0, 0.25)$ and $\alpha_{\sigma}$ is drawn from an exponential distribution with $\lambda = 1$. We construct a separate $\alpha_{v}$ for each unique $A_{id}$ where each $\alpha_{v} \sim \mathcal{N}(0, 1.0)$. These values are chosen because they help the MCMC sampling converge. $\beta$ and $\gamma$ follow the same construction except with different shapes for $\beta_{v}$ and $\gamma_{v}$. During the training of the hierarchical Bayesian model, we use 1000 warmup steps and generate an additional 1000 samples.

\subsection{Feature Agreement}
\label{appendix:featureagreement}

To demonstrate feature agreements, we calculate a weighted average to quantify the results of~\Cref{tab:feature-evaluation} in the context of the full feature set, using the correlations obtained from the Bayesian model as weights. The results are shown in~\Cref{tab:agreement-evaluation}.

\end{document}